\documentclass[aps,twocolumn,noshowpacs,superscriptaddress,floatfix]{revtex4-1}
\usepackage[table]{xcolor}
\usepackage{quantikz}
\usepackage{amsmath,amssymb}
\usepackage{mathrsfs}
\usepackage[justification=justified]{caption}
\usepackage{xcolor}
\usepackage{verbatim}
\usepackage[normalem]{ulem}
\usepackage[rightcaption]{sidecap}
\usepackage{graphicx}
\usepackage{subfigure}
\usepackage{caption}
\usepackage{multirow}
\usepackage{amsthm,epsfig,tikz}
\usepackage{float}
\usepackage{multirow}
\usepackage[colorlinks,citecolor=black,urlcolor=black,linkcolor =black,bookmarks=false,hypertexnames=true]{hyperref} 
\usepackage{relsize}
\usepackage{xcolor}

\usepackage{float}
\makeatletter
\let\newfloat\newfloat@ltx
\makeatother
\usepackage{algorithm}
\usepackage{algcompatible}

\renewcommand{\part}[2]{\frac{\partial #1}{\partial #2}}

\DeclareMathOperator*{\argmin}{arg\,min}

\begin{document}

\title{ {\large$\alpha$}QBoost: An Iteratively Weighted Adiabatic Trained Classifier}

\author{Salvatore Certo}
\email{scerto@deloitte.com}
\affiliation{Deloitte Consulting, LLP}

\author{Andrew Vlasic}
\affiliation{Deloitte Consulting, LLP}

\author{Daniel Beaulieu}
\affiliation{Deloitte Consulting, LLP}

\date{\today}

\begin{abstract}

A new implementation of an adiabatically-trained ensemble model is derived that shows significant improvements over classical methods.  In particular, empirical results of this new algorithm show that it offers not just higher performance, but also more stability with less classifiers, an attribute that is critically important in areas like explainability and speed-of-inference. In all, the empirical analysis displays that the algorithm can provide an increase in performance on unseen data by strengthening stability of the statistical model through further minimizing and balancing variance and bias, while decreasing the time to convergence over its predecessors. 

\end{abstract}

\maketitle

\section{Introduction} \label{sect:intro}
Leveraging quantum annealers, a new algorithm to adiabatically-train an ensemble method for classification is derived and denoted as {\large$\alpha$}QBoost. {\large$\alpha$}QBoost is empirically demonstrated to outperform classical methods while being more stable and compact than the classically trained models. Specifically, it is demonstrated that this algorithm can provide an increase in performance on unseen data, further minimizing bias/variance and strengthening stability of the statistical model.  A stable and compact ensemble model is critically important in areas like explainability - less complexity allows for easier deployment in highly regulated industries, especially when the ensmble model is comprised of simple weak classifiers like decision stumps.  

{\large$\alpha$}QBoost is extension of the annealing based classification algorithm, QBoost, that provides an improvement through faster convergence by re-weighting the linear and quadratic components of the problem.  By not including the penalty term to be included in the final model and instead employing an alternative regularization scheme, the optimal parameters can be found quicker. To strengthen the algorithm, a method to select classifiers given a target size of the ensemble is also proposed. The complete algorithm is then benchmarked against standard ensemble methods with two public data sets and one internal data set. With the focus on model performance and model stability, {\large$\alpha$}QBoost is demonstrated to provided considerable improvements over classical methods.


\section{Applications of Quantum Annealing} \label{sect:qa}
Quantum computers are theorized to solve problems currently intractable on classical computers, potentially offering exponential speedups to some of the hardest problems. Quantum annealers are processors specifically designed to solve combinatorial optimization problems, of which there are exciting applications in similar areas, especially in areas like unsupervised learning and training classical AI models \cite{schuman2019classical, neven2012qboost}.


Quantum annealing follows the adiabatic theorem, which states that if the time evolution is long enough, a quantum system will stay in the ground state of its Hamiltonian. As described in the foundation work in \cite{farhi2000quantum}, for a quantum system starting in a ground state of a Hamiltonian $H_b$ and evolving to the ground state of the problem Hamiltonian $H_p$, the system can be described mathematically at time $t$ out of the total run time $T$ as
\begin{equation*}
\mathcal{H}(t) = \big( 1-u(t) \big)\mathcal{H}_b + u(t)\mathcal{H}_p,
\end{equation*}
where $u(t) = t/T$.  Quantum annealing allows for interesting problems to be solved by formulating the problem into the Hamiltonian $H_p$.  At the end of the annealing process, the system is ideally in the ground state and the solution to our problem found.

The formulation of quantum annealing has quite a natural mapping to mathematical optimization problems, especially those of combinatorial optimization. Many of these problems are known to be NP-hard \cite{robinson2013introduction}. Several difficult problems, such as the Traveling Salesman Problem, have been known for decades.. Surprisingly, there have been applications outside of the well-known operations research problems, such neural networks which may be slightly reformulated as a Quadratic Unconstrained Binary Optimization (QUBO) problem \cite{sasdelli2021quantum} to solve for the parameters. While annealers are well suited for these combinatorial optimization problems with binary weights, many interesting problems also involve continuous of integer variables.  Therefore writing such problems into a QUBO requires a \textbf{binary encoding}. For completeness, consider binary encoding of a continuous variable within the unit interval, $[0,1]$, with $n$ binary variables denoted as $b_i$ for the $i^{th}$ variable, then the continuous variables has the form $\displaystyle \sum_{i=1}^n 2^{-i}b_i$.   

However, many general optimization algorithms, such as algorithms which use gradient descent, are naturally mapped to mathematical optimization. By reformulating the loss function of the model into a QUBO, the parameters of the model can be solved for explicitly without iterative methods like gradient descent. For example, the minimization formulation of linear regression and support vector machines are natural candidates to be reformulated as a QUBO with binary encoding. For instance, a prediction algorithm posed with finding weights with a $L^2$ loss function \cite{neven2012qboost} can be reformulated as a QUBO with a binary encoding. In fact, in general machine learning algorithms that utilize a $L^2$ loss function are candidates to be reformulated as a QUBO.  Interestingly, this reformulation may also be applied to solve ordinary differential equations and certain types of partial differential equations \cite{zanger2021quantum,garcia2022quantum}.

In the rest of this manuscript an extension of the adiabatically-trained classification algorithm QBoost \cite{neven2012qboost} is given. QBoost is a quantum implementation that determines the best combination of weak learners to reduce overfitting and increase performance on unseen data in ensemble methods.  It is based of off the pioneering algorithm Adaboost \cite{friedman2000additive}, which is the first classification algorithm to sequentially train weak learners at each iteration instead of an aggregation, like with random forests (see \cite{shalev2014understanding} for further information). An extension of QBoost is given by replacing the regularization term by an iterative weighting algorithm that determines the normalized weights for the linear and quadratic terms of the mathematical formulation.  This weighting helps to quickly find the best weak learners without over-training, resulting in well-balanced model with respect to the variance-bias trade-off.

\section{Binary Classification of a One Layer Weak Learner} \label{sect:qboost}
 Define $x_i \in \mathbb{R}^{N}$, for $N \in \mathbb{N}$, as either raw data, extracted features, mappings after a neural network or kernel, or an output from a weak classifier, all of which are known as a \textbf{general input pattern}. From $x_i$ define the associated labeled variable $y_i \in \{-1,1\}$. The goal is to understand weak binary classifiers of the general form $\tilde{y}_i = \mbox{sign}\big(w^T x_i +b )$ where $w \in \mathbb{R}^{N}$ is the weight and $b$ is the bias. To test from the known labels whether the classifier properly identified a label define $m(x,y,w,b) := y(w^T x+b)$, which is typically denoted as the \textit{margin}. The margin was given in some generality for various applications. 

For a collection of training data points $\big\{(x_i,y_i)\big\}_{i=1}^{S}$ to calculate the optimal weights the \textit{empirical risk} $\displaystyle \mathscr{L}(w,b) := \frac{1}{S} \sum_{s=1}^{S} l\Big( m(x_s,y_s,w,b) \Big)$, where $l$ is the loss function. The loss function may have different forms given the particular task. The classifier considered is a predictor that does not output a probability score. Given this predictor a potential loss function counts the missed classifications $l_{0-1}\Big( m(x,y,w,b) \Big) = \big( 1-\mbox{sign}( m(x,y,w,b) ) \big)/2$. However, this loss has been noted as NP-hard to minimize since $l_{0-1}$ is non-convex (\cite{feldman2012agnostic}), as well as not a good error to train with since there is no other information besides properly classified or not.

A more appropriate loss function is the square-loss $l_{ \mbox{sqr} }\Big( m(x,y,w,b) \Big) := \Big( m(x,y,w,b) - 1) \Big)^2$. Given this loss function, which is convex, the empirical risk may be utilized to derive the optimal weight and bias. However, it is well-known that deriving and utilizing the empirical risk may lead to an overfit statistical model. To adjust for this overfitting we can introduce a \textit{regularization term}, $\Lambda(w)$, that adds a `penalty' to the weights within the algorithm. Taking $\lambda \in \mathbb{R}_{\geq 0}$ as a parameter to adjust the strength of $\Lambda(w)$, the general regularization term is $\lambda \cdot \Lambda(w)$. For the training of this classifier $\displaystyle \Lambda(w) = ||w||_0 := \sum_{i=1}^{N} \delta_{>0}(w_i)$, where $\delta_{>0}$ is a step-function defined $\delta_{>0}(x) = 1$ if $x>0$ and $0$ otherwise. To compute the optimal classifier the goal is then to 
\begin{equation}\label{eq:formu}
    (w^*,b^*) = \argmin_{w,b} \Big\{ \mathscr{L}(w,b) + \lambda \cdot \Lambda(w)\Big\}.
\end{equation}

For the rest of this manuscript the general input patterns $\{x_i\}_{i=1}^{S}$ are cleaned and well-defined features, and the final strong classifier is an ensemble statistical model of the form  $\displaystyle \frac{1}{N}\sum_{i=1}^{N} w_i h_i(x_s)$ where $\{h_i\}_{i=1}^{N}$ is a collection of weak learners. 

Setting $b=0$ in Equation \ref{eq:formu}, the minimization problem may be written in a different form that gives deeper insight into how the collection of weak learners interact with each other. With $b=0$, observe Equation \ref{eq:formu} no has the form
\begin{equation}\label{eq:riskout}
    \begin{split}
        \argmin_{w} & \Bigg\{ \frac{1}{S} \sum_{s=1}^{S} \Big| \frac{1}{N} \sum_{i=1}^{N} w_i h_i( x_s ) - y_s\Big|^2 + \lambda \sum_{i=1}^{N} \delta_{>0}(w_i) \Bigg\}
        \\ = \argmin_{w} & \Bigg\{ \frac{1}{S} \sum_{s=1}^{S} \left( \frac{1}{N}\sum_{i=1}^{N} w_i h_i( x_s ) \right)^2 
        \\ & -\frac{1}{S} \sum_{s=1}^{S}\left(\frac{1}{N}\sum_{i=1}^{N} w_i h_i( x_s )  + y_s^2 \right)  + \lambda \sum_{i=1}^{N} \delta_{>0}(w_i) \Bigg\}
        \\ = \argmin_{w} & \Bigg\{ \frac{1}{N^2} \sum_{i=1}^{N}\sum_{j=1}^{N} w_iw_j\left( \frac{1}{S} \sum_{s=1}^{S} h_i( x_s )h_j( x_s ) \right) 
        \\ &  + \frac{1}{N}\sum_{i=1}^{N} w_i  \Big( \lambda -2 \frac{1}{S}\sum_{s=1}^{S} h_i( x_s )y_s \Big) \Bigg\}.
    \end{split}
\end{equation}

Note that $\displaystyle \frac{1}{S}\sum_{s=1}^{S}y_s^2 = 1$ for all sets $\{y_i\}_{i=1}^{S}$, and hence only adds a fixed scalar which does not affect the minimization and may therefore be dropped. Furthermore, notice that the term $\displaystyle \sum_{s=1}^{S} h_i( x_s )h_j( x_s )$ is the correlation term between the weak learners, and the term $\displaystyle\sum_{i=1}^{S}h_i( x_s )y_s$ is the correlation between a weak learner and the labels. 

Since the weight $w$ focuses on the importance of the weak learners this equivalent formulation is now a quadratic optimization minimization problem. Restricting each weight $w_i \in [0,1]$, to convert the quadratic optimization formulation into a QUBO each weight will be written as a binary encoding, where in particular, $\displaystyle w_i = \sum_{k=1}^{K} 2^{-k} w_{ik}$ with $w_{ik}$ as a binary variable. 

When each variable $w_i$ is binary encoded an additional binary variable, denoted as $w_i^{reg}$, is required for a proper count of nonzero terms in the regularization terms. To incorporate $w_i^{reg}$ into the formulation take $\kappa$ sufficiently large and define the regularization function as 

\begin{equation}
 R(w) = \sum_{i=1}^{N}\kappa w_i (1-w_i^{reg}) + \lambda w_i^{reg}.  
\end{equation}

For generality the regularization function is denoted as $R(w)$. Hence, when binary encoded variables are included for weighted classifiers the regularization function above will be implicitly noted and the original form in Equation \ref{eq:riskout} otherwise.  

For simplicity define
\begin{equation}\label{eq:q}
           \mathcal{Q}(w,\lambda) := \argmin_{w} \Bigg\{ \frac{1}{S} \sum_{s=1}^{S} \Big| \frac{1}{N}\sum_{i=1}^{N} w_i h_i( x_s ) - y_s\Big|^2 + R(w) \Bigg\}
\end{equation}

From this formulation into a QUBO the authors in \cite{neven2012qboost} derive an algorithm to find the optimal weights for the binary classifier. For a concise description of algorithm, define the general test metric function as $\mbox{TM}( \mathbf{y}_{true} , \mathbf{y}_{predicted} )$. Lastly, we extend the definition of $\mathcal{Q}(w,\lambda)$ by including the data $\{(x_i,y_i) \}_{i=1}^{S}$ with the notation $\mathcal{Q}(w,\lambda, \mathbf{x}, \mathbf{y})$, where $\mathbf{x} := \{ x_i \}_{i=1}^{ S }$ and $\mathbf{y} := \{ y_i \}_{i=1}^{ S }$. The algorithm is as follows.

\begin{algorithm}[H]
\caption{QBoost}\label{alg:qboost}
\raggedright\textbf{Input:} Training and validation set $\{(x_i,y_i) \}_{i=1}^{S}$, where the validation set is defined as $\mathbf{x}_{val} := \{ x_i \}_{i=1}^{ S_{val} }$ and $\mathbf{y}_{val} := \{ y_i \}_{i=1}^{ S_{val} }$, and the training set is defined as $\mathbf{x}_{train} := \{ x_i \}_{i=1}^{S_{train}}$ and $\mathbf{y}_{train} := \{ y_i \}_{i=1}^{S_{train} }$, and dictionary of weak classifiers $\{h_i\}$.

\begin{algorithmic}[1]
\STATE $d \gets \frac{1}{S_{train}}$, $T \gets 0$, $\mbox{Error}_{val} \gets \infty$, $\mathbf{w}^* \gets 0$
\STATE Initialize: empty strong classifier $\mathbf{H}_{w^*}(\cdot)$; storage for $Q$ candidates of weak learners $\{h_q\}$
\STATE Initialize $\lambda_{ \mbox{min} }$, $\lambda_{ \mbox{max} }$, and $\lambda_{ \mbox{step} }$
\STATE \textbf{repeat}
\STATE\hspace{\algorithmicindent} Optimize the members of $\{h_i\}$ according to the error $d$
\STATE\hspace{\algorithmicindent} From $\{h_i\}$ select $Q-T$ classifiers with the smallest errors weighted by $d$ and add them to $\{h_q\}$
    \STATE \hspace{\algorithmicindent} \textbf{for}  $\lambda = \lambda_{ \mbox{min} } : \lambda_{ \mbox{step} } : \lambda_{ \mbox{max} }$ \textbf{do}
\STATE\hspace{\algorithmicindent} \ \ \ $w^* \gets \mathcal{Q}(w,\lambda)$
\STATE\hspace{\algorithmicindent} \ \ \ $\mathbf{H}_{w^*}^{temp}(\cdot) \gets \mbox{sign}\left( \sum_{q=1}^{Q} w_i^* h_q(\cdot) \right)$
\STATE\hspace{\algorithmicindent} \ \ \ $\mbox{Error}_{temp} \gets \mbox{TM}\left(\mathbf{y}_{val}, \mathbf{H}_{w^*}^{temp}(\mathbf{x}_{val}) \right)$  
\STATE\hspace{\algorithmicindent} \ \ \ \textbf{if}  $\mbox{Error}_{val} > \mbox{Error}_{temp}$ \textbf{then}
\STATE\hspace{\algorithmicindent} \ \ \ \ \ \ $\mbox{Error}_{val} \gets \mbox{Error}_{temp}$
\STATE\hspace{\algorithmicindent} \ \ \ \ \ \ $T \gets  \sum_{i=1}^{N} \delta_{>0}(w_i)$
\STATE\hspace{\algorithmicindent} \ \ \ \ \ \ $\mathbf{H}_{w^*} \gets \mathbf{H}_{w^*}^{temp}$
\STATE\hspace{\algorithmicindent} \ \ \ \ \ \ $\mathbf{w}^* \gets w^*$
\STATE\hspace{\algorithmicindent} \ \ \ \textbf{end if}
\STATE\hspace{\algorithmicindent} \textbf{end for}
\STATE\hspace{\algorithmicindent} $d \gets d \cdot \left( \mathbf{y}_{train}\cdot \mathbf{H}_{w^*}(\mathbf{x}_{train}) - 1\right)^2 $
\STATE\hspace{\algorithmicindent} $\displaystyle d \gets \frac{d}{ \sum_{k=1}^{S_{train} } d_k }$
\STATE\hspace{\algorithmicindent} Delete from $\{h_q\}$ the $Q-T$ weak learners for which $w_i^*=0$
\STATE \textbf{end repeat} When $\mbox{Error}_{val}$ stops decreasing 
\end{algorithmic}
\end{algorithm}

Notice within Algorithm \ref{alg:qboost} the search for an optimal $\lambda$ is a naive linear grid search and this technique has potential to be implemented in a more optimal manner. Moreover, the range to consider the values of $\lambda$ is contingent on the size and quality of the training data.

\section{Hybrid Implementation of a Weighted Weak Learner}\label{sect:algo}
The minimization in Equation \ref{eq:riskout}, and hence the Algorithm \ref{alg:qboost}, is reliant on the regularization term to ensure the resulting model is not overfit. While incorporating a regularization term is a standard implementation for many statistical modeling algorithms, from the exploration of the formulation in Equation \ref{eq:riskout}, there is a more direct way to balance the correlation between each classifier and between the classifiers and the true labels. Unless otherwise noted, $\lambda = 0$ throughout the rest of the paper.

\begin{algorithm}[H]
\caption{Classifier Selection}\label{alg:bsw}
\raggedright\textbf{Input:} Training set $\{(x_i,y_i) \}_{i=1}^{S}$, where $\mathbf{x} := \{ x_i \}_{i=1}^{ S }$ and $\mathbf{y} := \{ y_i \}_{i=1}^{ S }$, and dictionary of weak classifiers $\{h_i\}$.
\begin{algorithmic}[1]
\STATE $a \gets 0$, $b \gets 1$, $\alpha \gets .5$, count $\gets 0$
\STATE Initialize: desired-count
\STATE $m^* \gets \mathcal{Q}'(w,\alpha, \mathbf{x}, \mathbf{y})$
\STATE count $\gets |\{h_i^*\}|$
\STATE \textbf{while} count $\neq$ desired-count \textbf{do}
    \STATE \hspace{\algorithmicindent} \textbf{if} count $>$ desired-count \textbf{then}
        \STATE \hspace{\algorithmicindent} \ \ \  $b \gets \alpha$
    \STATE \hspace{\algorithmicindent}  \textbf{else}
        \STATE \hspace{\algorithmicindent}  \ \ \ $a \gets \alpha$
    \STATE \hspace{\algorithmicindent} \textbf{end if}
    \STATE \hspace{\algorithmicindent} $\alpha \gets (a+b)/2$
    \STATE \hspace{\algorithmicindent} $m^* \gets \mathcal{Q}'(w,\alpha, \mathbf{x}, \mathbf{y})$
    \STATE \hspace{\algorithmicindent} count $\gets |\{h_i^*\}|$
\STATE \textbf{return} $m^*$
\end{algorithmic}
\end{algorithm}

The authors in \cite{mucke2022quantum} give a very rigorous formulation of a quantum annealing based feature selection algorithm where the correlation of the features and the true labels are weighted with $\alpha \in [0,1]$ and the correlation of the features against with one another are weighted $(1-\alpha)$. These weights are complimentary since are both non-negative and sum to $1$. After giving the specific number of desired features, the algorithm then utilizes the well-known binary search algorithm of a sorted array to find the proper value for $\alpha$. By balancing each correlations this bypasses the need to add in a regularization term. 

Recall from Equation \ref{eq:riskout} with the respective data $
\mathbf{x} := \{ x_i \}_{i=1}^{ S }$ and $\mathbf{y} := \{ y_i \}_{i=1}^{ S }$, the term $$\displaystyle \mbox{Cor}_1(w, \mathbf{x}, \mathbf{y}):= -2\frac{1}{N}\sum_{i=1}^{N} w_i \Big( \frac{1}{S} \sum_{s=1}^{S}h_i( x_s )y_s \Big)
$$ is the correlation between the weak learners and true labels, and the term $$\displaystyle \mbox{Cor}_2(w,\mathbf{x}) := \frac{1}{N^2}\sum_{i=1}^{N}\sum_{j=1}^{N} w_iw_j\Big(\frac{1}{S} \sum_{s=1}^{S} h_i( x_s )h_j( x_s ) \Big)$$ is the correlation between each of the weak learners. With these newly defined terms the minimization is re-written as
\begin{equation} \label{eq:quad}
\begin{split}
   \mathcal{Q}'(w, \alpha, \mathbf{x}, \mathbf{y}) := \argmin_{w} & \Big\{ \alpha \cdot \mbox{Cor}_1(w, \mathbf{x}, \mathbf{y})
   \\ & + (1-\alpha) \cdot \mbox{Cor}_2(w, \mathbf{x}) \Big\}.    
\end{split}
\end{equation}

Taking this new formulation and following the algorithm in \cite{mucke2022quantum} with fixed weak classifiers $\{h_i\}$, Algorithm \ref{alg:bsw} is an implementation to find $\alpha$ for the specified number of classifiers. Given the similarities of the model in \cite{mucke2022quantum} and Equation \ref{eq:quad}, using similar logic in the paper yields that the algorithm is guaranteed to converge to the specified number of classifiers. In general, this algorithm has the potential to be sub-optimal in finding the best number of classifiers when considering the error of the cross validation set, which may lead to either a biased model or a model with high-variance. One may test as to whether the model is sub-optimal and increase or decrease the number of classifiers, accordingly. The advantage, however, is that one may control the size of the classifier. Now taking Algorithm \ref{alg:bsw} as a subprocess a new derivation of QBoost may be presented.

 \begin{algorithm}[hbt]
 \caption{QBoost with Classifier Selection}\label{alg:aqb-bsw}
 \raggedright\textbf{Input:} Training and validation set $\{(x_i,y_i)  \}_{i=1}^{S}$, where the validation set is defined as $\mathbf{x}_{val} := \{  x_i \}_{i=1}^{ S_{val} }$ and $\mathbf{y}_{val} := \{ y_i \}_{i=1}^{ S_{val}  }$, and the training set is defined as $\mathbf{x}_{train} := \{ x_i  \}_{i=1}^{S_{train}}$ and $\mathbf{y}_{train} := \{ y_i \}_{i=1}^{S_{train}  }$, and dictionary of weak classifiers $\{h_i\}$.

 \begin{algorithmic}[1]
 \STATE $d \gets \frac{1}{S_{train}}$, $T \gets 0$, $\mbox{Error}_{val} \gets  \infty$, $\mathbf{w}^* \gets 0$
 \STATE Initialize: empty strong classifier $\mathbf{H}_{w^*}(\cdot)$; storage for $Q$ candidates of weak learners $\{h_q\}$
 \STATE \textbf{repeat}
 \STATE\hspace{\algorithmicindent} Optimize the members of $\{h_i\}$ according to the error $d$
 \STATE\hspace{\algorithmicindent} From $\{h_i\}$ select $Q-T$ classifiers with the smallest errors weighted by $d$ and add them to $\{h_q\}$
     \STATE \hspace{\algorithmicindent} $\mathbf{H}_{w^*} \gets$ Algorithm\ref{alg:bsw}$\Big(\{h_q\},\mathbf{y}_{train}, \mathbf{x}_{train}, \mathbf{y}_{val}, \mathbf{x}_{val} \Big)$
 \STATE\hspace{\algorithmicindent} $d \gets d \cdot \left(  \mathbf{y}_{train}\cdot \mathbf{H}_{w^*}(\mathbf{x}_{train}) - 1\right)^2 $
 \STATE\hspace{\algorithmicindent} $\displaystyle d \gets \frac{d}{ \sum_{k=1}^{S_{train} } d_k }$
 \STATE\hspace{\algorithmicindent} Delete from $\{h_q\}$ the $Q-T$ weak learners for which $w_i^*=0$
 \STATE \textbf{end repeat} When $\mbox{Error}_{val}$ stops decreasing 
 \end{algorithmic}
\end{algorithm}
\begin{table*}[t]
{ \begin{tabular}{ |c|c|c|c|c|}
 \hline \multicolumn{5}{|c|}{ \textbf{Comparison of Algorithms} } \\ \hline
 Test & Model & Breast Cancer & Heart Failure & Smart Factory\\ \noalign{\hrule height 2pt} 
 
 \multirow{3}{*}{Accuracy}  & AdaBoost & $95.61\%$ & $78.00\%$ & $78.67\%$ \\ \cline{2-5} 
  & {\large$\alpha$}QBoost & \textcolor{red}{$95.79\%$} & \textcolor{red}{$82.00\%$} & \textcolor{red}{$84.00\%$} \\ \cline{2-5}
  & Random Forest & $94.90\%$ & $81.66\%$ & \textcolor{red}{$84.00\%$} \\ \cline{2-5}
    & Gradient Boosted & $93.86\%$ & $78.66\%$ & $80.66\%$ \\ \noalign{\hrule height 2pt}
	
 \multirow{3}{*}{$f1$ Score}  & AdaBoost & $96.48\%$ & $63.86\%$ & $82.38\%$ \\ \cline{2-5}
  & {\large$\alpha$}QBoost & \textcolor{red}{$96.76\%$} & $68.45\%$ & $87.58\%$ \\ \cline{2-5}
  & Random Forest & $95.90\%$ & \textcolor{red}{$68.95\%$} & \textcolor{red}{$87.80\%$} \\ \cline{2-5}
    & Gradient Boosted & $95.10\%$ & $65.69\%$ & $83.80\%$ \\ \noalign{\hrule height 2pt}
 \multirow{3}{*}{Number of Classifiers}  & AdaBoost & $30$ & $30$ & $75$ \\ \cline{2-5}
& {\large$\alpha$}QBoost & \textcolor{red}{$15$} & \textcolor{red}{$5$} & \textcolor{red}{$40$} \\ \cline{2-5}
& Random Forest & $30$ & $30$ &$75$ \\ \cline{2-5}
& Gradient Boosted & $30$ & $30$ & $75$ \\ \noalign{\hrule height 2pt} 
\end{tabular} }
 \caption{Results for the 3 data sets.  Data was randomly sampled, trained, and tested 5 times and the results averaged. For the backend of {\large$\alpha$}QBoost, the breast cancer data Gurobi was employed, for the heart failure data D-Wave was employed, and for the Smart Factor data the simulated annealer was employed. }\label{tab:compare}
 \end{table*}

One may observe that the Classifier Selection algorithm in Algorithm \ref{alg:aqb-bsw} may be replaced with a classical gradient-free optimizer algorithm \cite{robinson2013introduction} with the goal of determining the optimal $\alpha$ which yields the optimal set of weights $w$. This is especially true with optimizers derived around quadratic optimization. There are a number of optimizers that include the well-known Bayesian optimizer \cite{frazier2018tutorial}, COBYLA \cite{robinson2013introduction}, and the Nelder-Mead method \cite{ozaki2017effective}. Denote $Opt$ as a general gradient-free optimizer. Just as in Algorithm \ref{alg:bsw}, the evaluation is conducted with the validation set.  With this observation the main algorithm is formulated in Algorithm \ref{alg:aqb}. 

\begin{algorithm}[hbt]
\caption{{\large$\alpha$}QBoost (Main)}\label{alg:aqb}
\raggedright\textbf{Input:} Training and validation set $\{(x_i,y_i) \}_{i=1}^{S}$, where the validation set is defined as $\mathbf{x}_{val} := \{ x_i \}_{i=1}^{ S_{val} }$ and $\mathbf{y}_{val} := \{ y_i \}_{i=1}^{ S_{val} }$, and the training set is defined as $\mathbf{x}_{train} := \{ x_i \}_{i=1}^{S_{train}}$ and $\mathbf{y}_{train} := \{ y_i \}_{i=1}^{S_{train} }$, and dictionary of weak classifiers $\{h_i\}$.

\begin{algorithmic}[1]
\STATE $d \gets \frac{1}{S_{train}}$, $T \gets 0$, $\mbox{Error}_{val} \gets \infty$, $\mathbf{w}^* \gets 0$
\STATE Initialize: empty strong classifier $\mathbf{H}_{w^*}(\cdot)$; storage for $Q$ candidates of weak learners $\{h_q\}$
\STATE \textbf{repeat}
\STATE\hspace{\algorithmicindent} Optimize the members of $\{h_i\}$ according to the error $d$
\STATE\hspace{\algorithmicindent} From $\{h_i\}$ select $Q-T$ classifiers with the smallest errors weighted by $d$ and add them to $\{h_q\}$
    \STATE \hspace{\algorithmicindent} $\mathbf{H}_{w^*} \gets \mbox{Opt}\Big(\{h_q\},\mathbf{y}_{train}, \mathbf{x}_{train}, \mathbf{y}_{val}, \mathbf{x}_{val} \Big)$
\STATE\hspace{\algorithmicindent} $d \gets d \cdot \left( \mathbf{y}_{train}\cdot \mathbf{H}_{w^*}(\mathbf{x}_{train}) - 1\right)^2 $
\STATE\hspace{\algorithmicindent} $\displaystyle d \gets \frac{d}{ \sum_{k=1}^{S_{train} } d_k }$
\STATE\hspace{\algorithmicindent} Delete from $\{h_q\}$ the $Q-T$ weak learners for where $w_i^*=0$
\STATE \textbf{end repeat} When $\mbox{Error}_{val}$ stops decreasing 
\end{algorithmic}
\end{algorithm}

Unlike Algorithm \ref{alg:qboost} where the range of $\lambda$ is unknown a priori, the range of $\alpha$ will always be in the unit interval, $[0,1]$. Furthermore, Algorithm \ref{alg:aqb} is robust enough to include a number of gradient-free optimizers and implementing Algorithm \ref{alg:aqb-bsw} is straightforward; Algorithm \ref{alg:aqb-bsw} is quite applicable when there is a restricted number of classifiers required for a specific task, for instance, when a model needs to be explainable.

A potential shortcoming, however, is the classifier computed in Algorithm \ref{alg:aqb} maybe calculated from a local minimum since the data utilized has an effect on convergence; this observation also holds for Algorithm \ref{alg:qboost}. A proposed mitigation follows the stochastic gradient descent to consistently feed the algorithm different data sets with each iteration. One potential implementation randomly draws the training set and validation set, each with fixed size throughout the algorithm, before each iteration of finding the optimal classifier with the pool of weak learners. However, this technique may not work quite well with large data sets. For large data sets one may preemptively draw $k$ different sets of training and validation records and at each iteration of optimization one of these sets is chosen at random. Overall, this is a method to bootstrap the data to train the algorithm. 


\section{Analysis of Model Performance}\label{sect:anal}
To display the efficacy of {\large$\alpha$}QBoost against Adaboost \cite{friedman2000additive}, as this algorithm was the basis for QBoost, different data sets containing binary classes are trained with both algorithms. {\large$\alpha$}QBoost is ran on the D-Wave quantum annealer \cite{Dwave} and D-Wave's simulated annealer, as well as on the Gurobi platform \cite{gurobi}, a classical solver. Furthermore, different gradient-free optimization algorithms are applied in the subprocess that computes the $\alpha$ parameter. The data for the analysis was taken from two open-source data sets Scikit-Learn \cite{pedregosa2011scikit} and the UCI repository \cite{Dua:2019}, and another data set from repository in Deloitte Smart Factory @ Wichita.

The Smart Factory data are images of housings for Smart Rover educational toys in partnership with Elenco Electronics. The data used in this research was created from Deloitte’s collaboration with the Smart Factory @ Wichita to help improve manufacturing of the modules, which contains a Raspberry Pi microcomputer, camera, body housings, and other accessories. The images analyzed by the models are of the Smart Rover body housings are of a collection of good Smart Rover housings, and defective Smart Rover housings. The defects in the Smart Rover body housing can appear on nearly any edge of housing of the body. The raw image data was encoded using Scikit-Learn and Imageio.\cite{Smart}. 

\subsection{Classifier Candidates}
As the {\large$\alpha$}QBoost algorithm proposed relies on an iterative process of proposing new weak learners weighted by an array $d$, it is necessary to ensure that the weak classifiers proposed are constructed in a close to optimal manner. A few simple methods have been proposed, like constructing a weak classifier for each feature, as well as for each combination of features multiplied by each other \cite{neven2012qboost}. From the data utilized, is was observed that applying such methods creates an unnecessary computational burden that can lead to increased iterations and a less efficient algorithm. Furthermore, is was also observed that the ensemble should allow for multiple classifiers built on the same feature in each iteration, but perhaps using a different loss function to decide the split. Even if the redundant classifier with the same splits are proposed, we believe this can aid in performance as it reduces the chance that the stronger weak classifiers are removed due to the randomness of the annealer.

Therefore, it is proposed to allow each classifier to be constructed using the best splits among multiple features, which is easily implemented using the `max features' attribute in Scikit-Learn.  Too little a value and the weak classifiers will be trained per each feature, too large and the pool of candidates will contain many instances of the same split. It was empirically observed that a value less than $\sqrt{N}$, where N is the number of features, leads to good performance.  Similarly, proposing $1.5-3$ times the number of candidates as the desired size of the ensemble also converges well.

Another option considered was using Adaboost's estimators as the first pool of candidates to filter, denoted as AdaStart-{\large$\alpha$}QBoost.  While there were noticeable increases in performance using this technique, it also led to increased instability and a larger number of classifiers in the final statistical model. However, this may be a result of the data sets (which were relatively small). Potential improvements to this algorithm are left to future research as to the scenarios in which this would be appropriate.

\begin{figure}[h]
    \centering
    \includegraphics[width=240px]{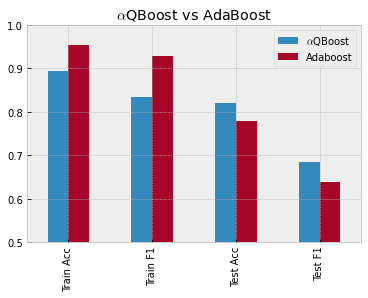}
    \caption{Example of higher generalization performance for {\large$\alpha$}QBoost on the Heart Failure data set. Adaboost frequently overtrained and suffered in test performance, whereas {\large$\alpha$}QBoost showed more stability with less classifiers.}
    \label{fig:sector_bands}
\end{figure}

\begin{figure}[h]
    \centering
    \includegraphics[width=240px]{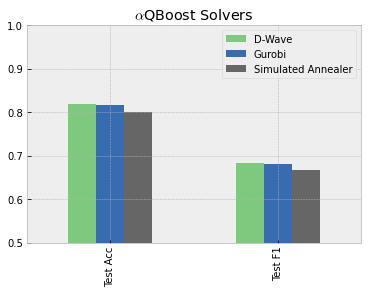}
    \caption{Comparison of solvers used to find the minimum energy of the QUBO on the heart failure data set. The results are all comparable, and all solvers arrived at a final model size of 5 weak learners, compared to the 30 weak learners computed by AdaBoost.}
    \label{fig:class}
\end{figure}

\subsection{Results}
We have found the performance of {\large$\alpha$}QBoost to be strong in comparison to Adaboost as well as other ensemble methods, resulting in equal or better performance with a more compact classifier.  In certain data sets like the UCI Heart Failure, the biggest increase in performance occurs when Adaboost fails to properly generalize to unseen data.  As shown in Figure \ref{fig:sector_bands} and Table \ref{tab:compare}, Adaboost clearly over-fits on the training data, suffering in the test set on both accuracy and the F1 score.  {\large$\alpha$}QBoost does not exhibit this behavior, instead sacrificing training performance for decreased variance and thus better performance on unseen data.  The variations in the optimizer can be attributed to the slightly stochastic optimizer in the inner loop, which we employed COBYLA to find the optimal value for $\alpha$.  

For a complete analysis, the backends of D-Wave, simulated annealer, and Gurobi are compared with the heart failure data set. These results are displayed in Figure \ref{fig:class}. While the results are similar, D-Wave id displayed to be the best backend. We posit that the annealer actually aids in this process by injecting slight randomness into the results that further prompts COBYLA to continue evaluating different values for $\alpha$.

Table \ref{tab:compare} shows a summary of the model performances across different data sets.  Rather than just compare to Adaboost, we also compared against additional ensemble methods like Random Forest and Gradient Boosted Trees.  For each data set, we randomly shuffled the samples and split them into train, validation, and test groups.  We ran the models 5 times and averaged the results.  The best performing model for each metric and data set is highlighted in red.  We found that {\large$\alpha$}QBoost was at least competitive, if not stronger, than the other standard models. Importantly, across all data sets {\large$\alpha$}QBoost resulted in a more compact classifier.


\section{Discussion}
The manuscript presented an extension of QBoost that directly balances variance and bias. This new algorithm, {\large$\alpha$}QBoost, is a robust one-layer weak learner algorithm that utilizes classical optimizers to compute the optimal weight, $\alpha$, which can increase speed-to-convergence compared to the original QBoost algorithm. Also, {\large$\alpha$}QBoost can be implemented to find the $N$ optimal combination of classifiers given the desired size of the final strong classifier. Furthermore, a method is proposed to derive the candidates of weak learners. 

From empirical analysis, it was observed that the classifier trained by {\large$\alpha$}QBoost was at worst competitive with classical algorithms, while consistently outperforming the statistical models with efficiency, stability, and compactness. To fully test the efficacy of this algorithm, with a large enough data set that may be split into training, testing, backtesting, and out-of-sample sets, it is suggested for further research to compare {\large$\alpha$}QBoost against other well-known and well-performing classifiers. 

\section{Disclaimer}
About Deloitte: Deloitte refers to one or more of Deloitte Touche Tohmatsu Limited (“DTTL”), its global network of member firms, and their related entities (collectively, the “Deloitte organization”). DTTL (also referred to as “Deloitte Global”) and each of its member firms and related entities are legally separate and independent entities, which cannot obligate or bind each other in respect of third parties. DTTL and each DTTL member firm and related entity is liable only for its own acts and omissions, and not those of each other. DTTL does not provide services to clients. Please see www.deloitte.com/about to learn more.

Deloitte is a leading global provider of audit and assurance, consulting, financial advisory, risk advisory, tax and related services. Our global network of member firms and related entities in more than 150 countries and territories (collectively, the “Deloitte organization”) serves four out of five Fortune Global 500® companies. Learn how Deloitte’s
approximately 330,000 people make an impact that matters at www.deloitte.com. 
This communication contains general information only, and none of Deloitte Touche Tohmatsu Limited (“DTTL”), its global network of member firms or their related entities (collectively, the “Deloitte organization”) is, by means of this communication, rendering professional advice or services. Before making any decision or taking any action that
may affect your finances or your business, you should consult a qualified professional adviser. No representations, warranties or undertakings (express or implied) are given as to the accuracy or completeness of the information in this communication, and none of DTTL, its member firms, related entities, employees or agents shall be liable or
responsible for any loss or damage whatsoever arising directly or indirectly in connection with any person relying on this communication. 
Copyright © 2022. For information contact Deloitte Global.

\bibliographystyle{unsrt}
\bibliography{qboost}
\end{document}